\documentclass[letterpaper, 10 pt, journal, twoside] {ieeetran}
\usepackage{generic}
\usepackage{cite}
\usepackage{amsmath,amssymb,amsfonts}
\usepackage{algorithmic}
\usepackage{graphicx}
\usepackage{textcomp}

\usepackage{multicol}
\usepackage{multirow}
\def\BibTeX{{\rm B\kern-.05em{\sc i\kern-.025em b}\kern-.08em
    T\kern-.1667em\lower.7ex\hbox{E}\kern-.125emX}}
\begin{document}
\title{A Flexible Piezoresistive/Self-Capacitive Hybrid Force and Proximity Sensor to Interface Collaborative Robots}
\author{Diogo Fonseca, Mohammad Safeea, and Pedro Neto
\thanks{This work was supported in part by the project PRODUTECH4S\&C (46102) by UE/FEDER through the program COMPETE 2020 and the Portuguese Foundation for Science and Technology (FCT): COBOTIS (PTDC/EME-EME/32595/2017) and UIDB/00285/2020. Paper no. TII-21-4994. (Corresponding author: Pedro Neto.) }
\thanks{Diogo Fonseca, Mohammad Safeea and Pedro Neto are with the Centre for Mechanical Engineering, Materials and Processes, University of Coimbra, 3004-504 Coimbra, Portugal (e-mail: diogo.fonseca@dem.uc.pt; ms@uc.pt; pedro.neto@dem.uc.pt).}}

\maketitle

\begin{abstract}
Force and proximity sensors are key in robotics, especially when applied in collaborative robots that interact physically or cognitively with humans in real unstructured environments. However, most existing sensors for use in robotics are limited by: 1) their scope, measuring single parameters/events and often requiring multiple types of sensors, 2) being expensive to manufacture, limiting their use to where they are strictly necessary and often compromising redundancy, and 3) have null or reduced physical flexibility, requiring further costs with adaptation to a variety of robot structures. This paper presents a novel mechanically flexible force and proximity hybrid sensor based on piezoresistive and self-capacitive phenomena. The sensor is inexpensive and easy to apply even on complex-shaped robot structures. The manufacturing process is described, including controlling circuits, mechanical design, and data acquisition. Experimental trials featuring the characterisation of the sensor were conducted, focusing on both force-electrical resistance and self-capacitive proximity response. The sensor's versatility, flexibility, thinness (1 mm thickness), accuracy (reduced drift) and repeatability demonstrated its applicability in several domains. Finally, the sensor was successfully applied in two distinct situations: hand guiding a robot (by touch commands), and human-robot collision avoidance (by proximity detection).
\end{abstract}


\section{Introduction}
\label{sec:introduction}
Advanced robots increasingly require more sensors to perceive their surroundings. Robot sensors are key to safe and intuitive human-robot interaction (HRI) as well as robot autonomy. It is often the case, however, that existing conventional sensors do not fit specific robots or application requirements. Conventional sensors are often unable to detect events of interest either due to insufficient scope, accuracy and/or robustness in working conditions. For example, vision sensors are constrained by light and suffer from occlusions. Other sensors may have excessive latency or limited repeatability. Frequently, researchers and engineers look for hybrid single sensor devices that can measure a wide scope of variables to detect events of interest, while being easy to install on a robot's structure (or in its surroundings). These sensors are, most often, not available. Existing devices frequently require costly alterations to reach hybrid, integrated solutions and they are usually not mechanically flexible, which further hampers their integration on structures with complex geometries.

The robotics and artificial intelligence markets are rapidly growing and sensors have a leading role in their success \cite{ref1}. Sensors become particularly important in a context where robots operate in more unstructured environments, often requiring interaction with humans. Tactile/force and proximity sensing are much-desired features for robot safety and interaction \cite{ref2}. The ISO/TS 15066 standard for collaborative robot safety significantly depends on the capability to detect the presence of humans or objects within the robot's surroundings, mainly for speed and separation monitoring \cite{ref3}. In this context, hybrid force and proximity sensors, particularly ones that are easy to install on pre-existing robot systems, become extremely desirable. 

This paper proposes a novel mechanically flexible and self-adhesive piezoresistive/self-capacitive hybrid force and proximity sensor that is both inexpensive and easy to apply even on complex-shaped surfaces. The development process is described here step by step, including design, manufacturing, complementary circuitry, and signal processing. Experiments featuring sensor characterisation were conducted, focusing on both its piezoresistive and self-capacitive operation modes. Finally, the proposed sensor is demonstrated in two distinct situations: 1) hand guiding a robot (by touch commands), and 2) collision avoidance (by proximity detection). From this work resulted the following contributions:

\begin{figure*}[!t]
	\centerline{\includegraphics[width=0.73\textheight]{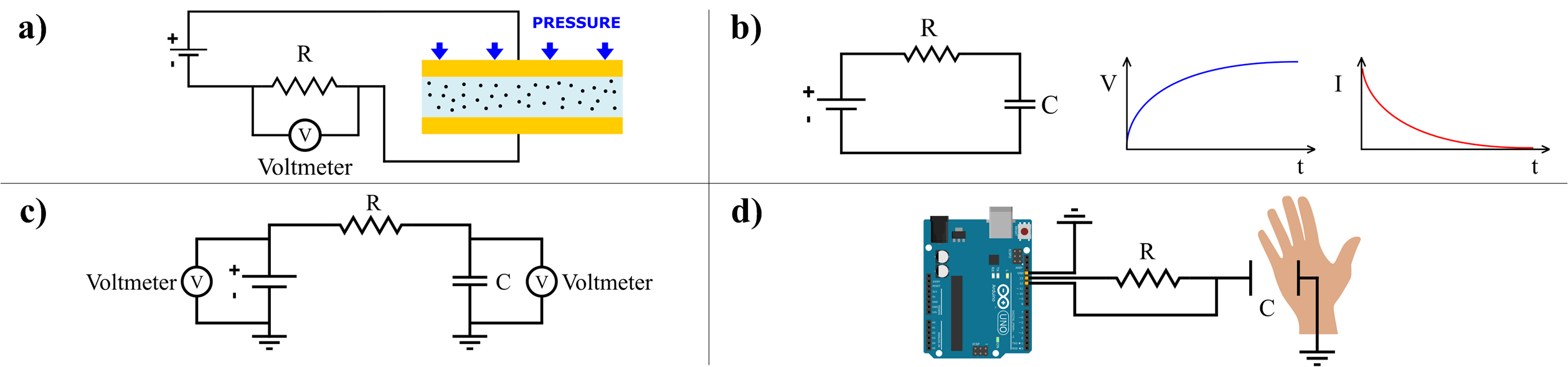}}
	\caption{a) Compression of a semiconductive polymer, b) Resistor-Capacitor (R-C) circuit, c) R-C circuit with common ground, d) Self-capacitive proximity sensing circuit.}
	\label{fig:fig1}
\end{figure*}

\begin{enumerate}
	\item A novel piezoresistive/self-capacitive hybrid force and proximity sensor; 
	\item The sensor's structure is mechanically flexible and self-adhesive making it easy to install even on complex-shaped structures;
	\item The sensor and its complementary circuitry are easy to manufacture, highly scalable and inexpensive (the manufacturing process is explained here step by step);
	\item The sensor provides reasonable accuracy and repeatability concerning force and proximity measurements. The presence of a human hand is detected at distances up to 100 mm away from an 8 cm$^2$ sensor while the minimum detectable force is 0.5 N and the single point repeatability is about 11\%;
	\item The sensor's versatility, flexibility and accuracy demonstrated its potential as an interface for collaborative robots.
\end{enumerate}

\subsection{Related Work}
Dual-mode sensors (capacitive and inductive) have demonstrated the ability to measure tactile pressure of up to 330 kPa and sense proximity with a good spatial resolution (3 mm) at distances of up to 150 mm \cite{ref4}. More recently, a dual proximity sensor combining the principles of inductive and capacitive proximity sensing was demonstrated, although with no pressure sensitivity \cite{ref5}. Our proposed sensor requires significantly less complex signal processing and complementary hardware than those presented in both \cite{ref4} and \cite{ref5}, albeit featuring a lower proximity resolution (30 mm). A soft Force Sensing Resistor (FSR) sensor capable of three-axial force measurements has been presented with interesting results, but is incapable of proximity detection \cite{ref6}. Recent research has also shown the potential of using commercially available FSRs for proximity sensing, taking advantage of the electrodes already present in the FSR to perform capacitive proximity sensing \cite{ref7}. A method for eliminating blind areas in piezoresistive sensor arrays has been presented, with moderate success \cite{ref8}. A tactile sensor for smart prosthetics, based on giant magneto-impedance material has shown very interesting results \cite{ref9}. Its scalability however may be limited, and proximity detection is not featured. Smart conductive textiles are increasingly used in wearable electronics. Proximity measurements have been demonstrated by measuring the electrostatic capacity in conductive flexible fabrics \cite{ref10}. Recently, conductive elastomers were studied to address the challenges of sensing deformable soft structures \cite{ref11}. Printable and stretchable elastic conductors, with initial conductivity of up to 738 Scm$^{-1}$, have been shown to endure up to 215\% strain while maintaining a conductivity of 182 Scm$^{-1}$ \cite{ref12}.	Programmable sensor networks have been integrated modularly in the form of a tape (flexible electronics substrate) \cite{ref13}, achieving considerably higher precision than our proposed sensor. Our solution is, however, over 200 times less expensive (by area), easier to manufacture, and easier to scale. In a recent study, contact forces and proximity were measured by arrays of proximity sensors in flexible printed circuit boards embedded in silicone \cite{ref14}. A robot skin composed of modularized hexagonally shaped cells is proposed in \cite{ref15}. Each cell includes a set of sensors reporting vibrations (3D accelerometer), pressure (three capacitive force sensors), pre-touch (optical proximity sensor), and temperature (two temperature sensors). This skin has a wider scope than our proposed solution (which cannot measure neither temperature nor acceleration). Once again, both \cite{ref14} and \cite{ref15} achieve higher resolutions particularly in the proximity detection, but with limited mechanical flexibility and manufacturing costs two to three orders of magnitude higher than the sensor proposed in this study. Electrostatic ultrathin flexible pressure sensors were successfully demonstrated in health-monitoring applications such as breathing monitoring \cite{ref16} and finger manipulation monitoring \cite{ref17}. Flexible pressure sensors based on organic-transistor-driven active matrices have shown great potential in the past two decades. Features ranging from optical transparency and bending-insensitivity \cite{ref18} to simultaneous pressure and temperature mapping \cite{ref19} have been demonstrated with promising results. All these solutions are significantly more complex to manufacture, although low-cost production should be achievable through high-volume production processes.

\subsection{Background and Design Principals}
When a conductive material is deformed, electrical resistance across it changes. Volume resistivity is often considered constant, making electrical resistance solely a function of geometric dimensions. This is the underlying principle behind various sensors, such as resistive strain gauges \cite{ref20}. However, semiconductive polymer composites such as Caplinq\textquotesingle s Linqstat or 3M\textquotesingle s Velostat behave differently. They feature a suspension of conductive filler particles randomly dispersed in a nonconductive polymeric matrix. Under strain, those particles move closer together in random movements known as Micro-Brownian Motion \cite{ref21}, Fig. \ref{fig:fig1} a). This effect increases the number of contacting particles, creating more paths for electricity to flow through. Simultaneously, non-contacting particles become, on average, separated by shorter distances, elevating the number of electrons tunnelling through the non-conductive matrix between them \cite{ref22}. Both mechanisms contribute to a great reduction of the material\textquotesingle s electrical resistivity under strain.

Proximity sensors are usually based on optical, ultrasonic, or capacitive technology. We were aiming for a thin, highly scalable, and flexible sensor so both ultrasonic and optical solutions were considered impractical. The self-capacitance phenomenon was deemed adequate to meet our design criteria. In Fig. \ref{fig:fig1} b) a simple RC circuit is shown where, after an initial transient period, a steady-state is reached. The potential across the capacitor's terminals opposes the power supply voltage and, at that point, no electric current flows through the circuit. A similar circuit may be considered, Fig. \ref{fig:fig1} c), where after a certain time, both the power supply and the capacitor will also tend to reach symmetric potentials. The resistor R, wired in series, increases the transient period (note that the time constant is directly proportional to the circuit\textquotesingle s resistance). Consider now a third circuit, Fig. \ref{fig:fig1} d), in which electrostatic energy is stored not inside a capacitor, but in an electric field created between an electrode and a nearby conductive surface, such as a human hand. Two digital pins in a microcontroller board are connected to the ends of the circuit. The first pin is turned high (+5V, in case of an Arduino Uno), and the second pin is set to input (without pullup). A counter variable is incremented in a loop until the potential reaching the second pin triggers a high state (at around 2.5V). The final value of this counter variable will be directly proportional to the circuit\textquotesingle s time constant and, in turn, to the circuit\textquotesingle s capacitance (considering R to be constant). Higher resistance values increase the transient period, allowing for more sensitive measurements at the expense of longer response times. Capacitance is inversely proportional to the distance between the hand and the electrode, but it can no longer be determined by the parallel capacitor formula as a plethora of other variables become involved (skin conductivity, air humidity, interference from the surrounding environment, etc.). Voltage measurements are also dependent on the reference ground values, which may vary. This method is relatively imprecise for distance measuring, but it enables reliable human presence detection, is easy to implement, and requires only basic hardware, while also being highly scalable.

\section{Piezoresistive/Self-Capacitive Sensor}

\begin{figure}[!t]\centering
	\includegraphics[width=1\columnwidth]{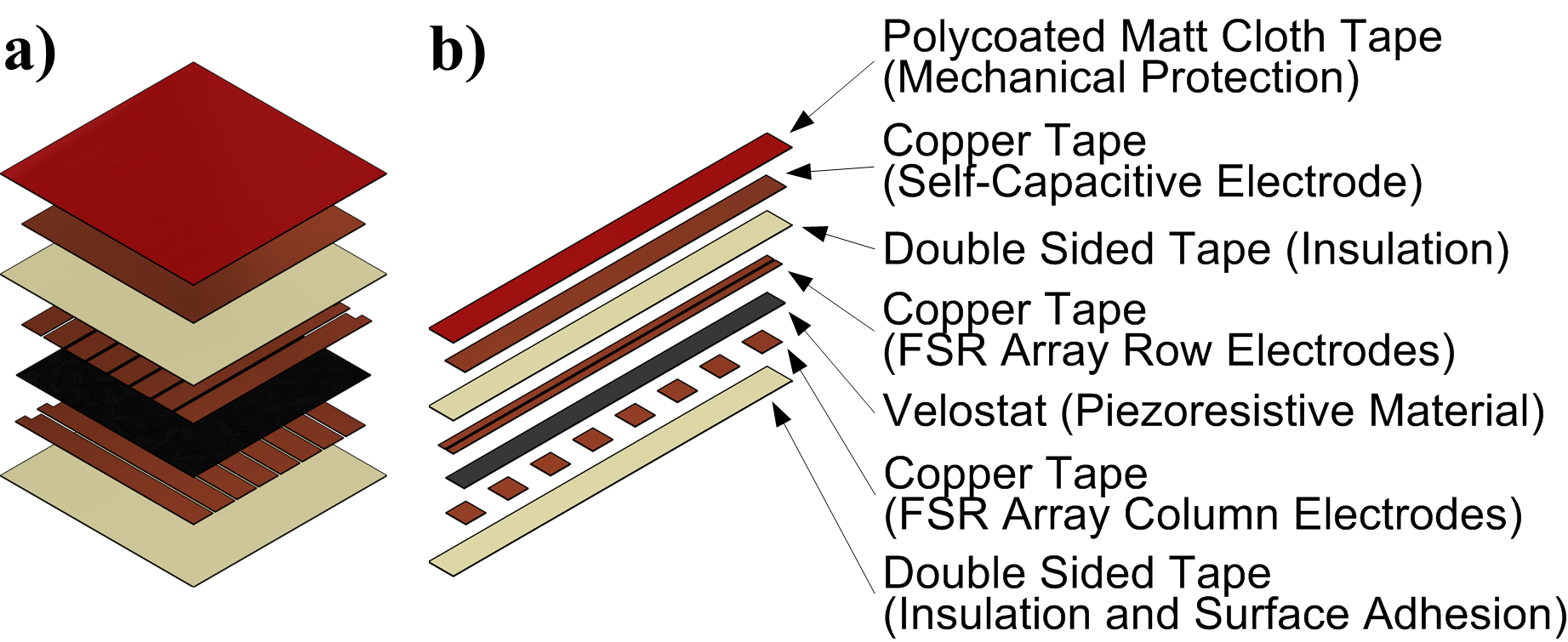}\centering
	
	\caption{Exploded view of the sensor\textquotesingle s elements. a) 64 prexel variant. b) 16 prexel variant.} \label{fig:fig2}
\end{figure}

\subsection{Design}

The development of the proposed piezoresistive/self-capacitive hybrid sensor was driven by the following design goals:

\begin{enumerate}
	\item Easy and inexpensive manufacturability, with minimal specialized tooling; 
	\item Good tactile resolution;
	\item Force sensing range of up to 15 N;
	\item Human presence detection at distances up to 100 mm;
	\item Mechanical flexibility and reduced thickness.
\end{enumerate}

Two sensor variants were developed: one featuring a FSR array with 16 prexels (ie. unitary pressure-sensitive elements) of 48 mm$^2$ each, arranged in a 2 $\times$ 8 configuration, and a second higher resolution one, featuring 64 prexels of 81 mm$^2$ each, arranged in an 8 $\times$ 8 configuration. The blind area of each sensor can be practically eliminated by design (ensuring short distances between adjacent prexels) or by application of the connected structure-based method presented by L. Wang \cite{ref8}. We made our sensors compatible with Wang\textquotesingle s method by designing a continuous layer of piezoresistive material (Fig. \ref{fig:fig2}). The area of each prexel was made similar to that of the tip of a human finger, ensuring adequate spatial resolution for tactile robot control (8.8 mm periodicity in the 8 $\times$ 8 version). A final design consideration was the area of the self-capacitive electrode to guarantee effective proximity detection. The maximum detectable distance should be in the order of the size of the electrode. We implemented an 8 cm$^2$ electrode in the 64 prexel sensor, aiming for our goal of 100 mm proximity detection range. 
\begin{figure}[!t]\centering
	\includegraphics[width=1\columnwidth]{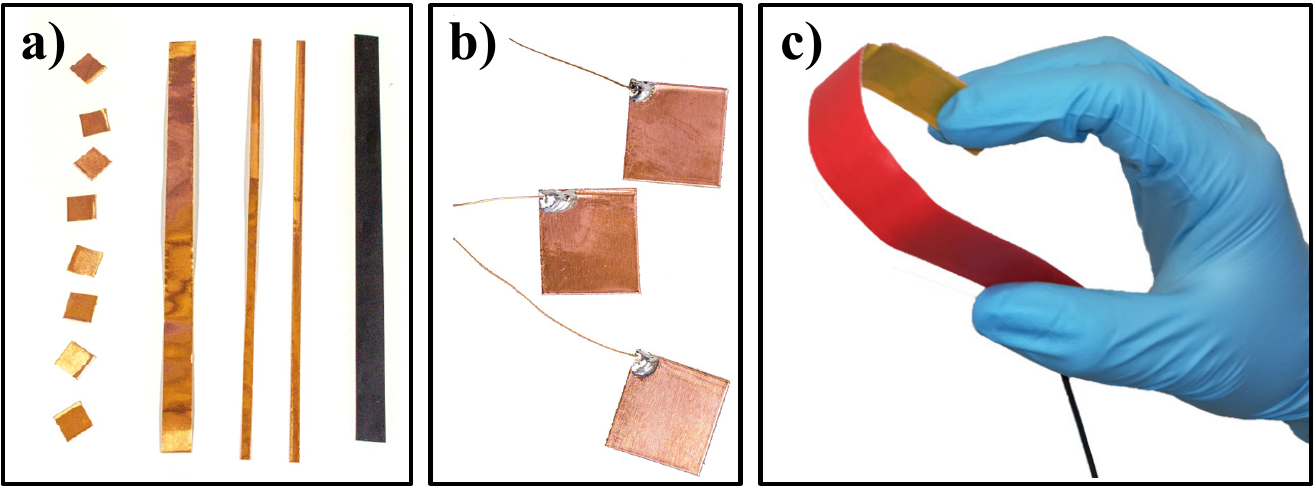}\centering
	
	\caption{a) Copper electrodes and Velostat strip, b) Detail of solder connections, c) Assembled 16 prexel sensor.} \label{fig:fig3}
\end{figure}

\subsection{Manufacturing}
The manufacturing process is composed of nine steps, including assembly of all the sensor\textquotesingle s elements, Fig. \ref{fig:fig2}. These steps are presented here for the 16 prexel variant, being analogous for the 64 prexel one:

\begin{figure*}
	\includegraphics[width=0.74\textheight]{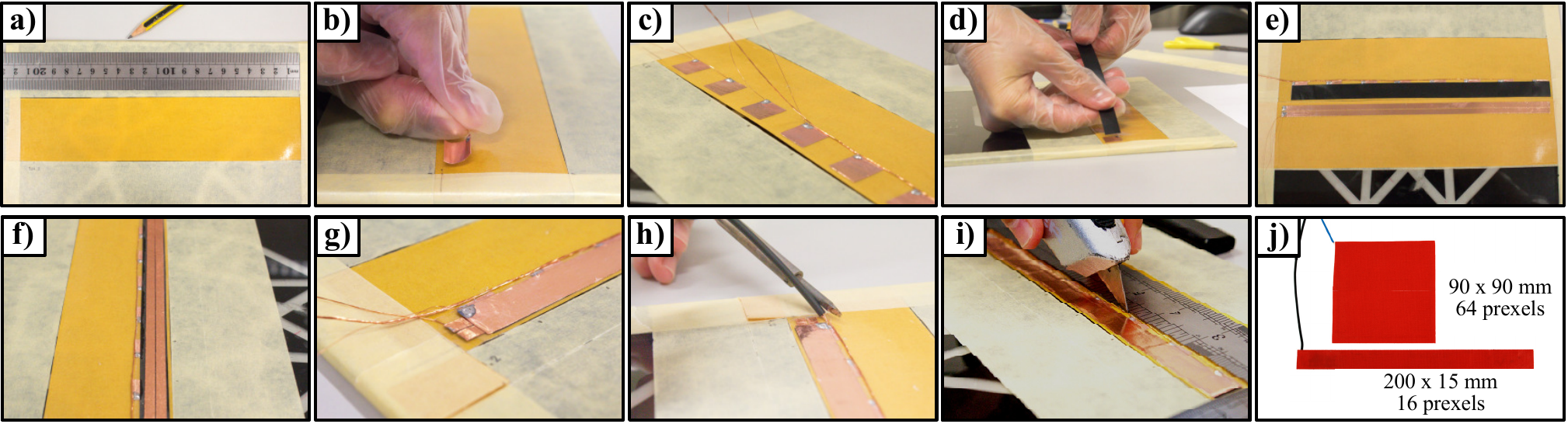}
	
	\caption{a) Base layer, b) Glueing the column electrodes of the FSR array, c) Wire management (twisted and glued to the base layer beneath), d) Semiconductive polymer installation, e) Lower and upper halves of the FSR array, before assembly, f) Completed FSR array, g) Self-capacitive electrode, h) Heat shrink tube installation, i) Cutting the sensor assembly to final dimensions, j) Completed Sensors.
		\label{fig:fig4}}
\end{figure*}

\begin{enumerate}
	\item Copper tape and Velostat pieces were cut to form the sensor\textquotesingle s internal elements. These include 8 prexel column electrodes, 2 prexel row electrodes, 1 Velostat layer and 1 self-capacitive copper electrode, Fig. \ref{fig:fig3} a);
	\item Enamelled copper wires were soldered to each copper electrode, Fig. \ref{fig:fig3} b). It is important to avoid the flow of solder to areas that will be in direct contact with the semiconductive polymer. Solder build-ups outside those areas should also be kept to the bare minimum, as they will create pressure concentration zones and promote layer separation, decreasing the sensor\textquotesingle s low-pressure sensitivity through a sharp increase in the electrical contact resistance between layers. Finally, all copper parts were thoroughly cleaned with isopropyl alcohol, both before and after the soldering process;
	\item To make the sensor\textquotesingle s base layer, a strip of double-sided tape was unrolled on top of a flat surface with its adhesive side up and held in place with strips of painter\textquotesingle s tape. We used a pencil to mark the measurements, Fig. \ref{fig:fig4} a);
	\item The 8 column electrodes were glued to the base layer, Fig. \ref{fig:fig4} b). Their enamelled wires were also twisted and glued to the base layer beneath, Fig. \ref{fig:fig4} c). All wires must be inside the sensor\textquotesingle s design dimensions;
	\item The Velostat layer was laid on top of the 8 column electrodes, Fig. \ref{fig:fig4} d). It is essential to ensure proper insertion at the first attempt. Unsticking and relocating the polymer may result in contamination of it\textquotesingle s surface with adhesive residues;
	\item A second strip of double-sided tape was placed on the workbench with its uncovered sticky side up. This will be the sensor\textquotesingle s middle insulating layer, separating the FSR array\textquotesingle s row electrodes from the self-capacitive electrode. After appropriate measurements, both array\textquotesingle s row electrodes were glued in place, Fig. \ref{fig:fig4} e);
	\item The middle insulating layer, with the 2 previously installed electrodes, was placed on top of the Velostat layer and the lower cover of the adhesive was removed, Fig. \ref{fig:fig4} f);
	\item The self-capacitive electrode was then installed, and all wires were arranged as flat as possible to reduce the risk of layer delamination, Fig. \ref{fig:fig4} g). A heat shrink tube was installed to provide further insulation and surface protection to the enamelled wires, Fig. \ref{fig:fig4} h);
	\item The sensor was cut to shape and covered with its final layer of polycoated matt cloth tape which will provide surface protection, Fig. \ref{fig:fig4} i). The fully assembled prototype is shown in Fig. \ref{fig:fig4} j).
\end{enumerate}

\subsection{Data Acquisition System}
The Data Acquisition System (DAQ) is composed of three main components:
\begin{enumerate}
	\item A custom-made Arduino shield;
	\item The Arduino itself, which runs the sensor's transfer functions and communicates with an external serial client;
	\item An external computer, which receives sensor data and interfaces with the robot's controller.
\end{enumerate}

\begin{figure}[!t]\centering
	\includegraphics[width=1\columnwidth]{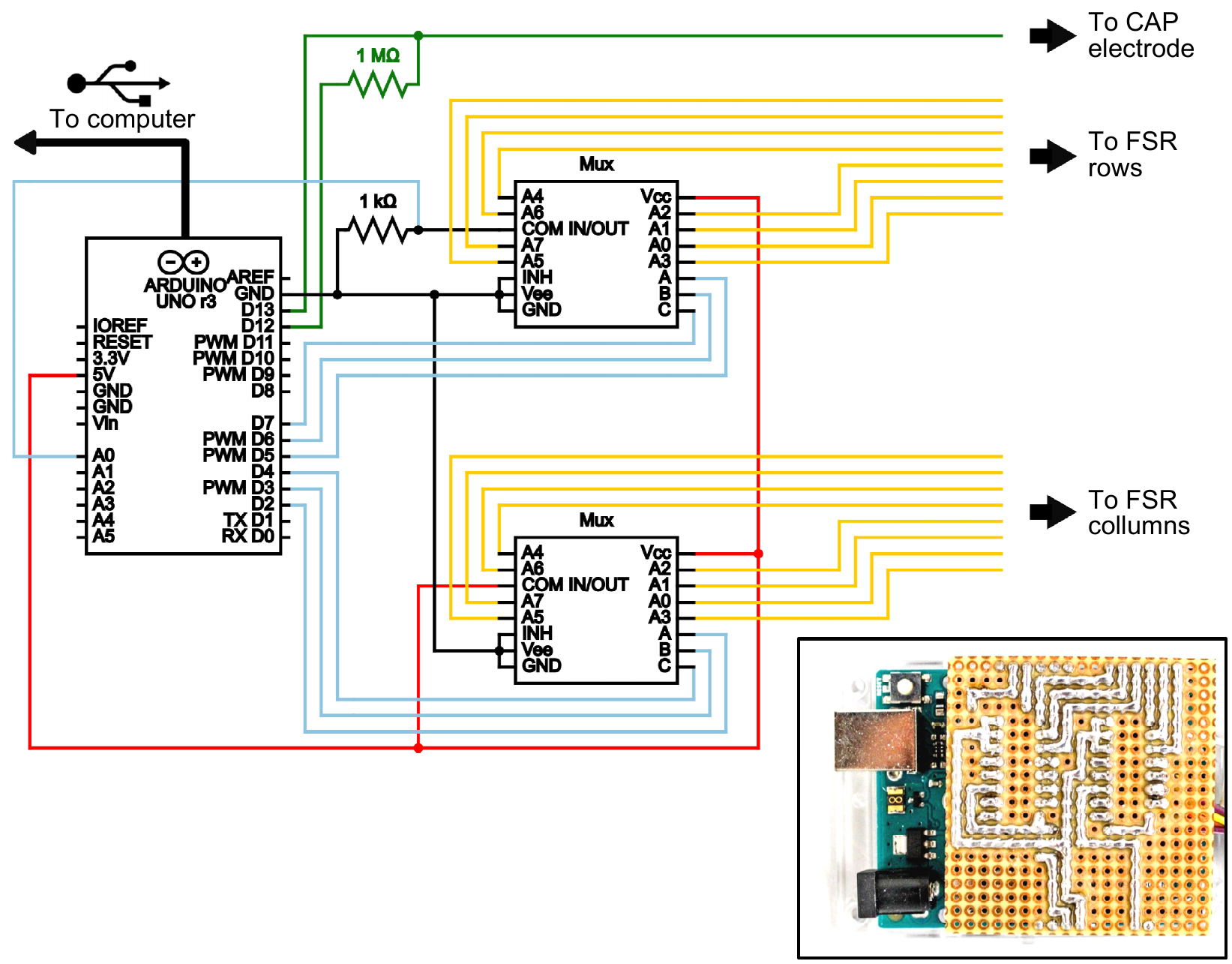}\centering
	\caption{DAQ circuit diagram (center) and assembled DAQ system (right).} \label{fig:fig5}
\end{figure}

The shield, circuit in Fig. \ref{fig:fig5}, features two 3-bit multiplexers (STMicroelectronics M74HC4051) to address the column-row pairs in the FSR array. This enables a maximum array size of $2^3$ $\times$ $2^3=64$ prexels. Higher bit multiplexers will enable exponentially larger sensor arrays. We used a voltage divider circuit with a reference resistor of 1 k$\Omega$ to measure the electrical resistance across each prexel. A 1 M$\Omega$ resistor, connected between the sending and receiving pins of the proximity sensing circuit, allows for adequate proximity sensitivity. 

The Arduino (Uno) was used to: 1) establish the connection to the serial client, 2) measure the transient period of the self-capacitive circuit (and apply a transfer function to determine the approximate distance in mm), 3) address both multiplexers to connect to each prexel in the FSR array, 4) measure voltage drop across each prexel, calculate corresponding resistance and apply a transfer function to determine the force being applied, 5) send all values to the serial client, and 6) return to point 2).

Self-capacitance measurements were performed using the algorithm developed by Paul Bagder and released to the public domain in the form of the Arduino "CapacitiveSensor" Library (version v0.4) \cite{ref23}. In short, a counter variable is first set to zero. Then, the send pin is set to high (+5V) and the counter is incremented in a loop until the self-capacitive electrode\textquotesingle s potential reaches +2.5V, causing the receive pin\textquotesingle s state to turn high. This charge cycle is then repeated a predefined $n$ number of times to smooth out high frequency noise (we found $n=70$ to be appropriate), with the counter variable being further incremented in each cycle. The increment frequency of the counter variable was determined to be $f=270$ kHz, so each unitary counter value corresponds to approximately 3.7 $\mu$s. The counter\textquotesingle s final value is then used as a relative measure of transient time and thus a relative measure of capacitance, according to:

\begin{equation}
C=\frac{-\Delta t}{R \times ln(1/2)}=\frac{-counter}{n \times f \times R \times ln(1/2)} \label{eq: rel_m}
\end{equation}
where $C$ is the capacitance, $\Delta t$ is the transient period, $R$ is the value of the resistor connected in series with the self-capacitive electrode, $n$ is the number of charge cycles and $f$ is the frequency at which the counter variable is incremented during each charge cycle. Every time someone approaches the sensor, its self-capacitance increases, resulting in higher transient times and thus higher counter values. 

Finally, an external computer was used to receive real-time data from the Arduino using a MATLAB-based platform, allowing for simultaneous real-time robot motion control using the KUKA Sunrise Toolbox (KST) \cite{ref24}.

\section{Sensor Characterization and Experiments}

A comprehensive theoretical approach for FSR modelling was proposed in \cite{ref25}. However, it requires advanced material analysis techniques which may be inaccessible to some researchers. For that reason, we propose an empirical approach. The characterisation of the self-capacitive proximity response was achieved using a semi-empirical model. Table \ref{tab:results_sg_alccuracy-1} details some of the main specifications and characteristics of the sensor.

\begin{table}
	\caption{Sensor specifications and characteristics}
	\label{tab:results_sg_alccuracy-1}
	\setlength{\tabcolsep}{2pt}
	\begin{tabular}{|p{165pt}|p{80pt}|}
		\hline
		Parameter& 
		Value \\
		\hline
		Size of tactile elements& 
		4x12 mm$^2$ \\
		Thickness& 
		1 mm \\
		Minimal detectable force (force sensing)& 
		0.5 N \\
		Single point repeatability (at 8.1 N) & 11.3\% \\
		Static loading (relative to measured force) & 6.1\% (after 3 hours) \\
		Drift (3 hours at a constant force of 8.1 N) & 5.8\% per log(min) \\
		Presence detection range (proximity sensing) & 0$-$100 mm \\
		Hysteresis (force sensing) & 15\% \\
		Rise time (force sensing) & 34.5 ms \\
		Delay time (force sensing) & 4.8 ms \\
		Sensitivity (force sensing) & 1.38x10$^{-4}$ S/N \\
		Precision (force sensing) & 7\% \\
		Resolution (force sensing) & 0.07$-$2.3 N \\
		\hline
	\end{tabular}
\end{table}

\subsection{FSR Empirical Model}

\subsubsection{Force-Electrical Resistance}

\begin{figure}[!t]\centering
	\includegraphics[width=1\columnwidth]{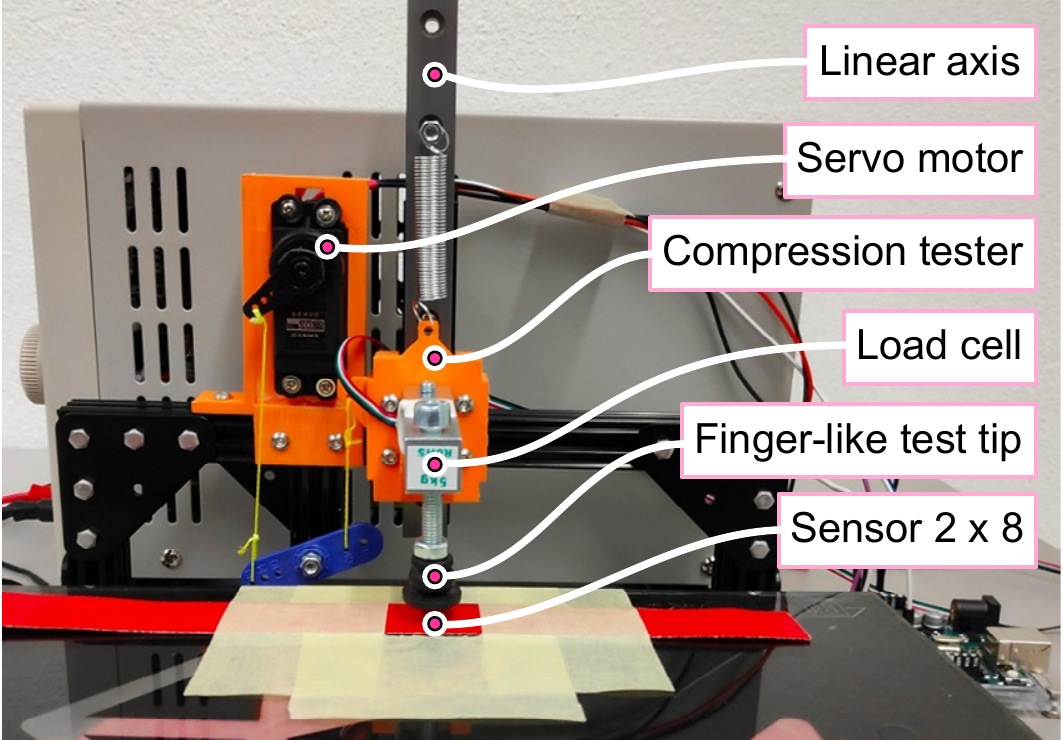}\centering
	
	\caption{Experimental setup of the force-electrical resistance compression/hysteresis test.} \label{fig:fig6}
\end{figure}

\begin{figure}[!t]\centering
	\includegraphics[width=1\columnwidth]{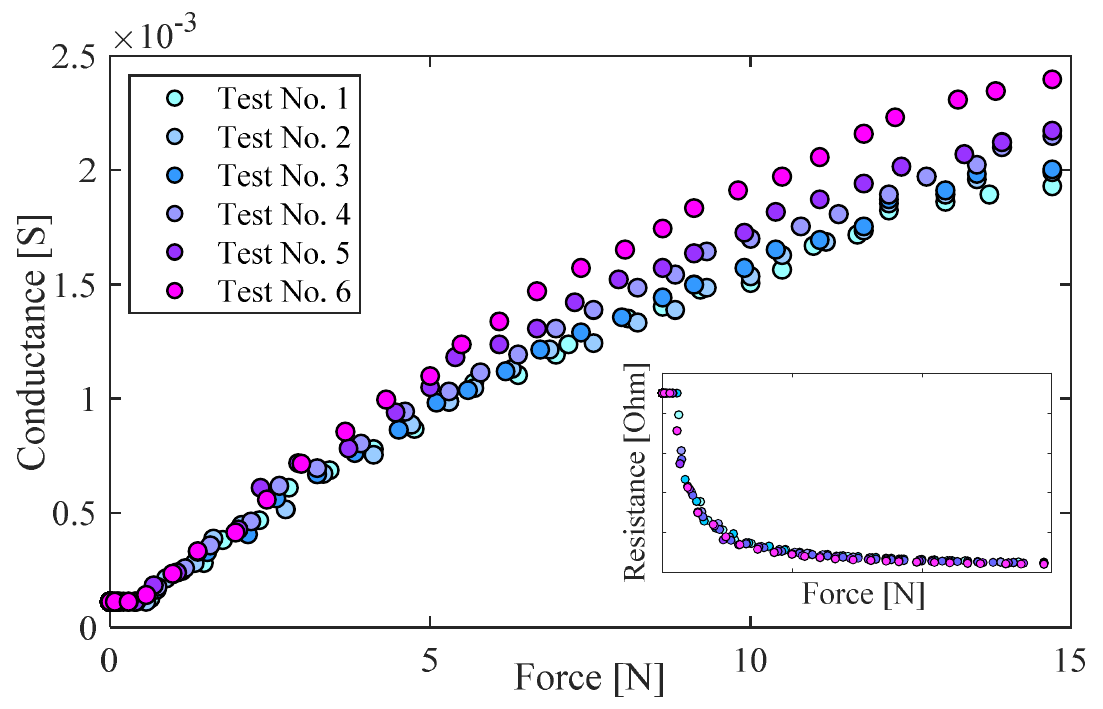}\centering
	\caption{Measured electrical conductance and electrical resistance.} 
	\label{fig:fig7}
\end{figure}

\begin{figure}[!t]\centering
	\includegraphics[width=1\columnwidth]{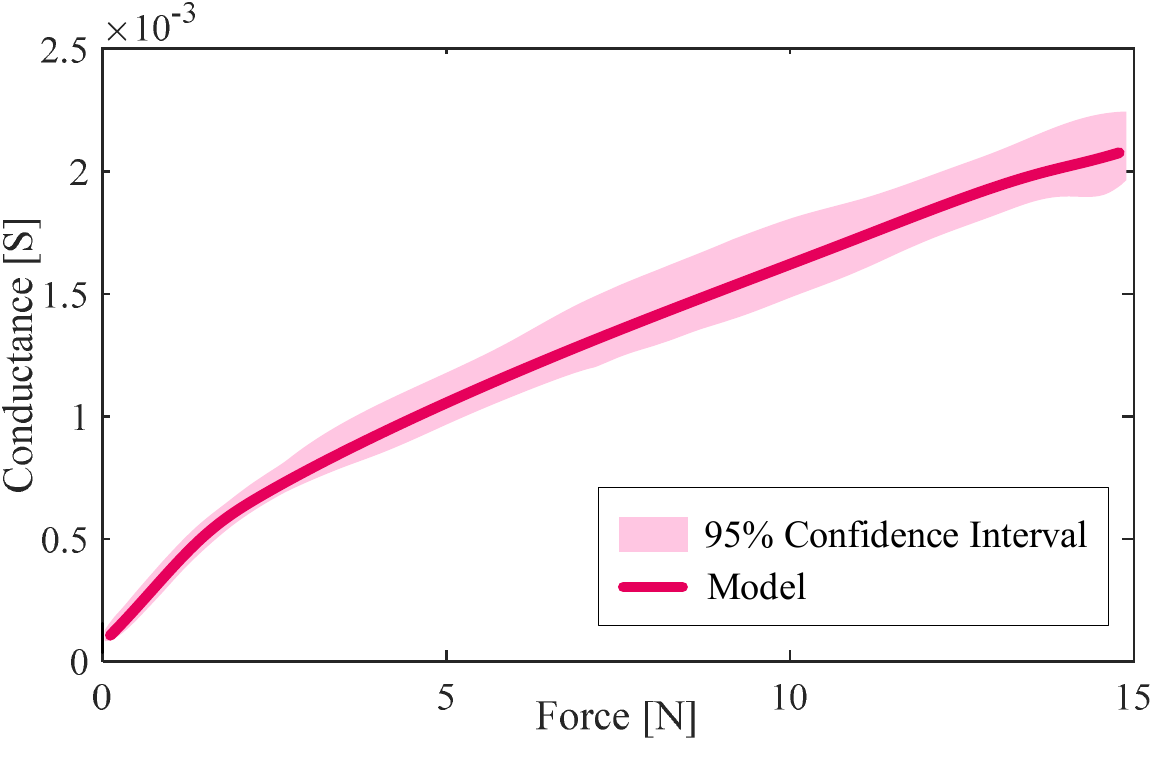}\centering
	\caption{Force-conductance FSR model with 95\% confidence intervals.}
	\label{fig:fig8}
\end{figure}

The force-electrical resistance relationship was evaluated using a compression test stand designed for the purpose, Fig. \ref{fig:fig6}. It features a rubber pressure test tip designed to closely emulate the area and pressure distribution of a human finger touching the sensor. In the first experiment, the sensor was subjected to forces ranging from 0 to 15 N in step increments of approximately 0.4 N. The resulting values of electrical resistance were recorded and plotted in Fig. \ref{fig:fig7}. The test was repeated 6 times at an average room temperature of 25ºC. Operation at higher temperatures will increase the sensor\textquotesingle s force sensitivity, due to the softening of the piezoresistive polymer layer. This phenomenon was not modelled in this study, and no significant variation was observed in the 18ºC to 25ºC range. The sensor is expected to endure operation at temperatures ranging from -50ºC up to 80ºC, according to specifications provided by the manufacturers of both the piezoresistive layer (3M) and the adhesive tapes forming the sensor\textquotesingle s structure (Advance Tapes).

Force-conductance data show an approximately linear behaviour, making linear regressions a good solution to obtain simple and computationally inexpensive empirical models. There are, however, some pointwise exceptions consistent to all 6 recorded data sets. These deviations occur mainly due to the layer delamination caused by manufacturing imperfections inside the sensor. To account for these deviations, high degree polynomial regressions can be used to better fit the force-conductance data. Choosing between linear and higher degree regressions requires a balance between computational cost and model accuracy. Also, as expected, the sensor proved to be inaccurate when subjected to forces lower than 0.5 N, mainly due to the delamination phenomena. Small pockets of air get between the semiconductive polymer and the electrodes inside the sensor, creating an unpredictable increase of contact electrical resistance. We did not take these low force/conductance values into consideration in our model, as they represent an unpredictable behaviour and were considered non-significant data. For our specific model, a high degree polynomial was fitted to the average force-conductance values across all 6 tests, and the standard deviations for each point were calculated. Confidence intervals of 95\% were also calculated, assuming multiple measures of the same force follow a normal distribution, Fig. \ref{fig:fig8}.

\subsubsection{Signal Drift}
\begin{figure}[!t]\centering
	\includegraphics[width=1\columnwidth]{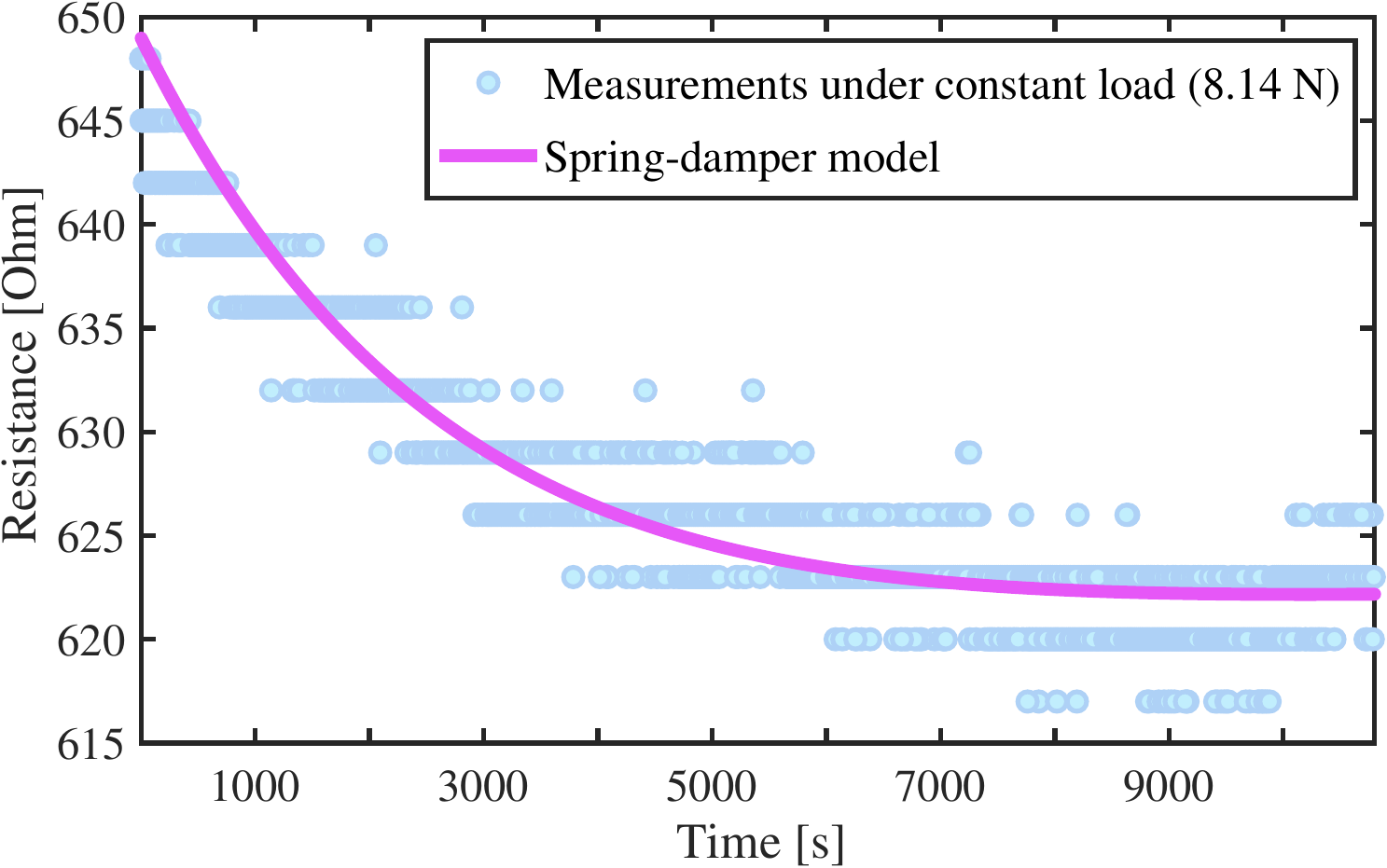}\centering
	
	\caption{Signal drift during a 3 hour test, and spring-damper model.} \label{fig:fig9}
\end{figure}

\begin{figure}[!t]\centering
	\includegraphics[width=1\columnwidth]{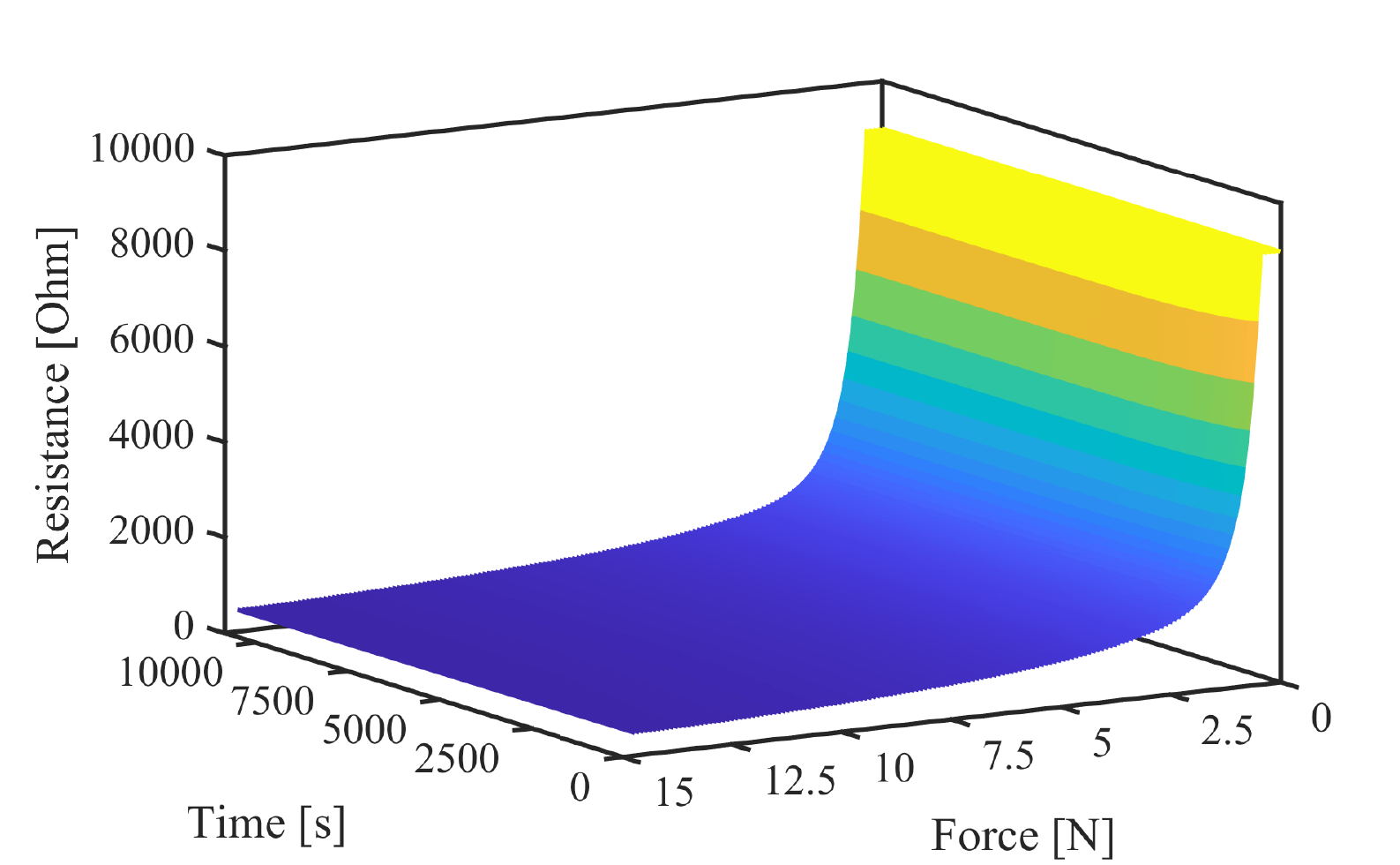}\centering
	
	\caption{Three-dimensional FSR model considering stress relaxation.} \label{fig:fig10}
\end{figure}

In a second experiment, a constant force was applied during a 3-hour test, to determine the sensor's signal drift. The data resulting from this test show an asymptotic decrease in electrical resistance over time, Fig. \ref{fig:fig9}. Velostat's electrical resistance is mostly a function of material strain. This explains why the graph in Fig. \ref{fig:fig9} shows a similar behaviour to the stress-time graph of a viscoelastic material undergoing stress relaxation. Such behaviour can be modelled by a critically damped spring-damper system, whose position could be determined by:
\begin{equation}
x(t)=(x_{0}+(\dot{x}_{0}+\omega_{n}x_{0})\times t)\times e^{-\omega_{n}t}\label{eq:Sistema Cr=0000EDtico}
\end{equation}
where $x_{0}$ is the system\textquotesingle s initial position, $\dot{x}_{0}$ is the system\textquotesingle s initial velocity and $\omega_{n}$ is the system\textquotesingle s natural angular frequency. Adapting (\ref{eq:Sistema Cr=0000EDtico}) to model the signal drift of our sensor, we have:
\begin{equation}
R(t)=(\Delta R+(A+B\times\Delta R)\times t)\times e^{-Bt}+R_{0}-\Delta R\label{eq:Modelo Critico}
\end{equation}
where $R_{0}$ is the initial electrical resistance, $\Delta R$ is the difference between the initial and final electrical resistance measured, $A$ and $B$ are parameters that can be determined by a numeric method. The previous force-resistance and resistance-time models were then combined, generating a complete three-dimensional empirical model, Fig. \ref{fig:fig10}.

\begin{figure}[!t]\centering
	\includegraphics[width=1\columnwidth]{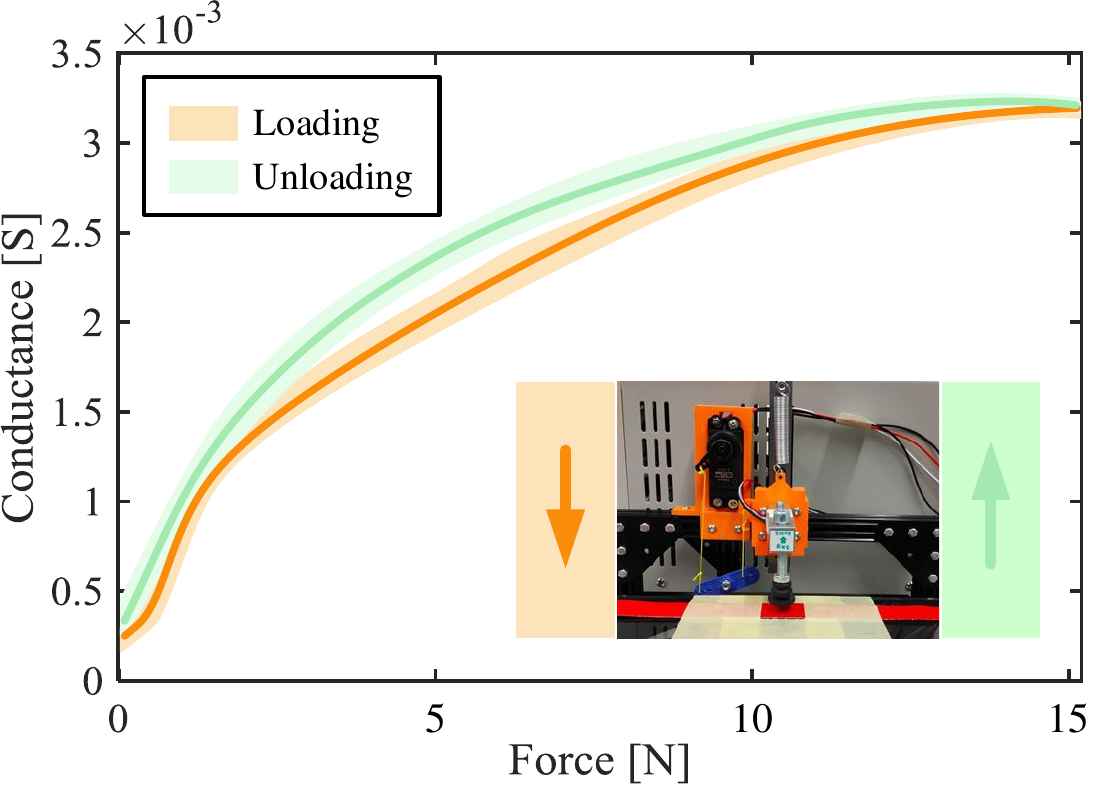}\centering
	
	\caption{Hysteresis curve during loading and unloading (16 cycles).} \label{fig:fig10-1}
\end{figure}

\begin{figure}[!t]\centering
	\includegraphics[width=1\columnwidth]{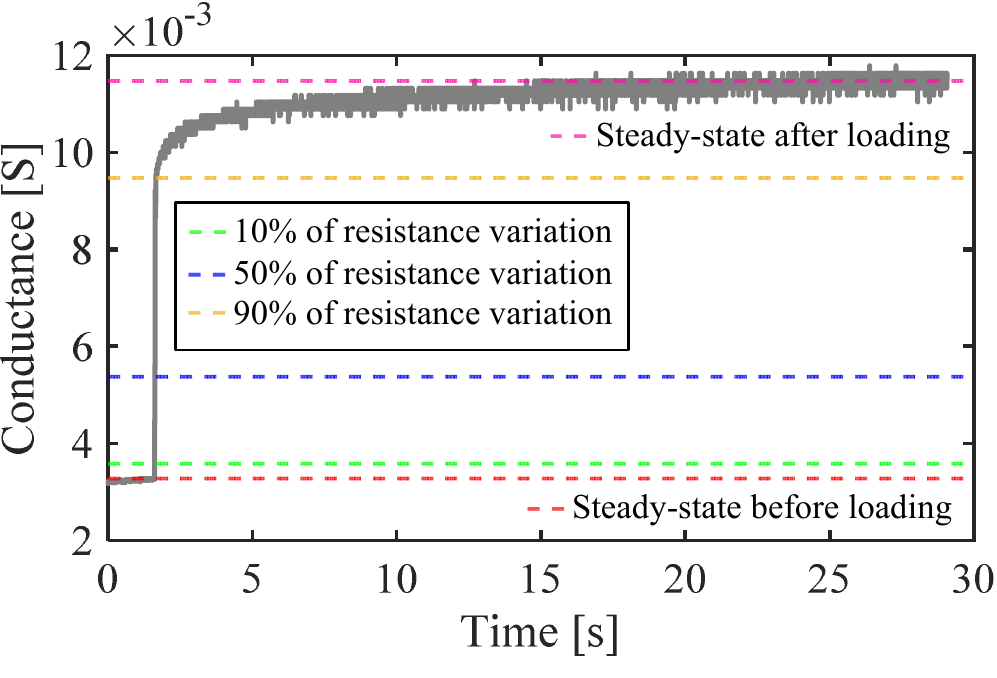}\centering
	
	\caption{Dynamic step response to point load.} \label{fig:fig10-2}
\end{figure}

\subsection{FSR Dynamic Response}

\subsubsection{Hysteresis}

Using the same setup shown in Fig. \ref{fig:fig6}, 16 load-unload cycles were performed with forces ranging from 0 to 15 N at a rate of approximately 0.7 N/s, Fig. \ref{fig:fig10-1}. Hysteresis errors ranged from 12 to 17\%, averaging 15\% across all 16 cycles. These values are in line with those of commercially available FSR's, suggesting that our sensor's additional layers and capacitive electrode do not significantly affect performance.

\subsubsection{Step Response}
We have also evaluated the sensor's response to a tactile step load. The inherent elasticity of either a human finger or the rubber test-tip in the experimental setup would overshadow the sensor's response delay, so a rigid 1.1 kg wooden sphere was used instead the rubber test-tip. The sphere was rested on top of one prexel, with a nylon cable supporting part of its weight. Then, the cable was cut so that the whole weight of the sphere was applied to the sensor. Measured conductance-time data are shown in Fig. \ref{fig:fig10-2}. As would be expected from any FSR, conductance values in Fig. \ref{fig:fig10-2} are significantly higher than the ones previously observed in Fig. \ref{fig:fig7}. This is due to the higher pressure created from by the sphere's small contact area. The delay time (from 0\% to 50\% of the steady state value) is 4.8 ms and the rise time (from 10\% to 90\% of the steady state value) is 34.5 ms, Table \ref{tab:results_sg_alccuracy-1}.

\subsection{Self-Capacitive Proximity Response}
The 16 prexel (15 $\times$ 200 mm$^2$) sensor proved reasonably effective at detecting the presence of a human hand up to 50 mm away, while the larger 64 prexel variant (90 $\times$ 90 mm$^2$) could detect presence at 100 mm. Being hybrid sensors, featuring both FSR and self-capacitive modes, they are able to distinguish between the presence and the touch of a human hand in their surroundings. As previously discussed, proximity readings start as a dimensionless time value, the counter. A semi-empirical model will directly correlate this value with distance values, in millimetres. Three different experiments were conducted.

In the first experiment, both sensors were placed on top of a non-conductive surface without any objects in close proximity. Proximity data were then recorded. This allowed us to determine a base value for each sensor. Results were an average counter value of 1486, with a standard deviation of 29 for the 16 prexel sensor, and 1610 with a standard deviation of 9.6 for the 64 prexel variant.

\begin{figure}[!t]\centering
	\includegraphics[width=1\columnwidth]{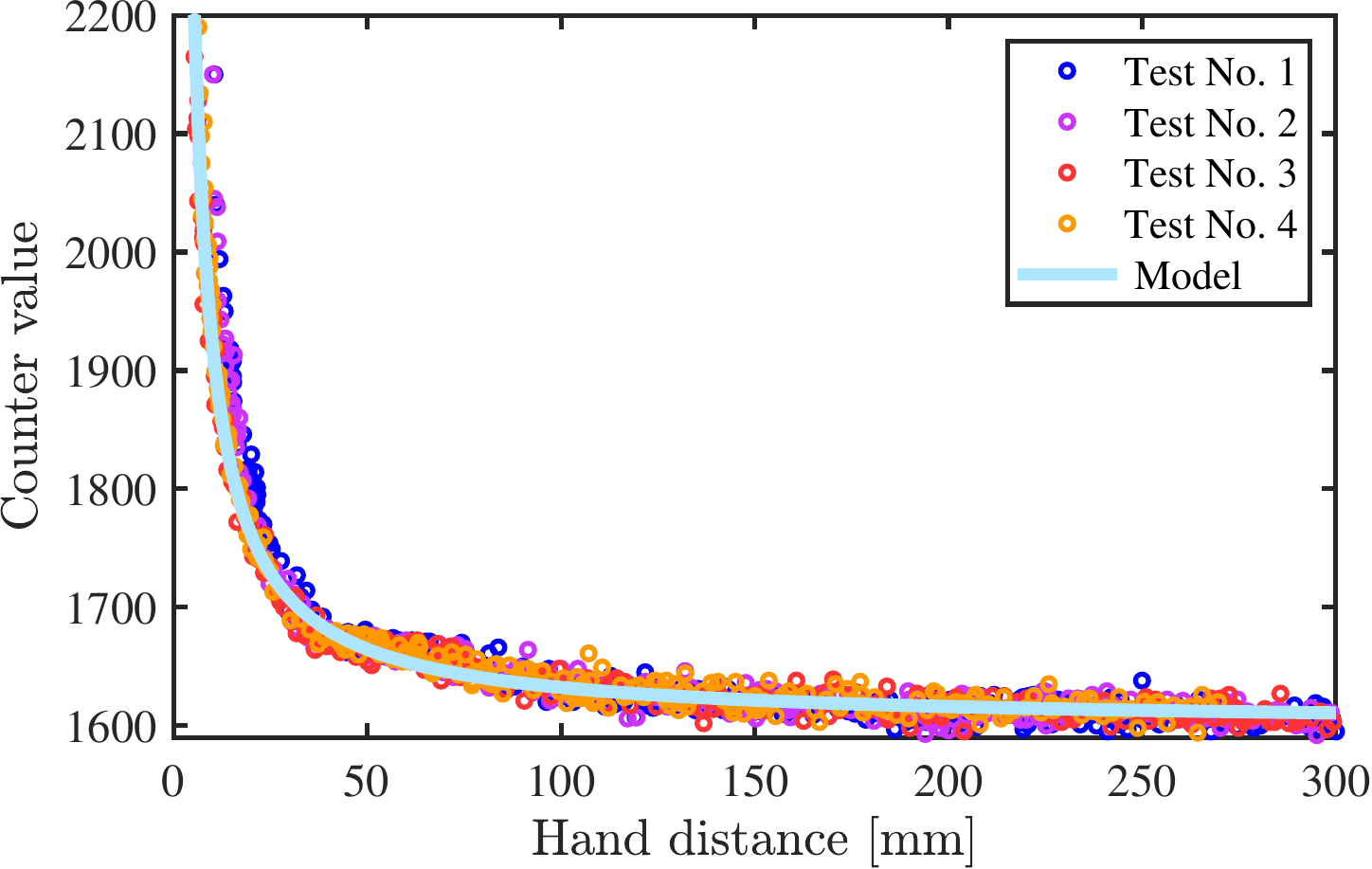}\centering
	\caption{Second experiment for the detection of a hand at different distances from the sensor. The hand is reliably detected at distances up to 100 mm.}
	\label{fig:fig11}
\end{figure}

In the second experiment, a human hand gradually approached the 64 prexel sensor until physical contact was established. Real-time hand distance was recorded using an electromagnetic tracker (Polhemus Liberty), which is accurate within +- 0.5 mm. The whole process was repeated 4 times. Sensor counts plotted against tracker readings can be seen in Fig. \ref{fig:fig11}. A 6th-order Butterworth low pass filter with a cutoff frequency of 1 Hz proved effective at smoothing the signal, although higher cutoff frequencies may be required depending on specific application. The voltage across a charging capacitor can be determined by:
\begin{equation}
V(t)=V_\infty\left(1-\exp\left(\frac{-t}{RC}\right)\right)\label{eq: Charging Capacitor}
\end{equation}
where $V(t)$ is the voltage across the capacitor, $V_\infty$ is the applied voltage, $R$ and $C$ are the circuit\textquotesingle s resistance and capacitance, respectively.

From (\ref{eq: Charging Capacitor}) it can be shown that the time necessary to reach a certain voltage across a capacitor is linearly dependent on its capacitance (ie. the time constant $\tau\propto C$). In a parallel plate capacitor, C is inversely proportional to the distance between electrodes. It is reasonable to assume that our sensor\textquotesingle s self-capacitance is also inversely proportional to the hand\textquotesingle s distance from it. Finally, the value of the counter variable is proportional to time, leading to:
\begin{equation}
counter=n \times f \times \Delta t =\frac{a}{x}+b\label{eq: proximity model}
\end{equation}
where $n$ is the number of charge cycles, $f$ is the incrementing frequency of counter, $\Delta t$ is the transient period, $x$ is the distance from the hand to the sensor, $b$ is the previously determined counter base value, and $a$ is a calibration parameter which can be determined experimentally using sensor readings over known distances. Experiments showed $a\approx2b$ to be a good first approximation. Fig. \ref{fig:fig11} shows the application of (\ref{eq: proximity model}) to experimental data.

The third and final experiment was similar to the second, but with several objects instead of a human hand approaching the sensor: a small anodized aluminium rod, an aluminium calliper, a steel hammer and a 500 ml plastic water bottle (full). The sensor could not reliably detect the presence of any of these objects. This represents a limitation of the proposed sensor, but some interesting side-effects arise: the sensor becomes capable of easily distinguishing the touch of a human hand from the touch of any of these objects, by comparing values between its piezoresistive and self-capacitive modes of operation.

\section{Application in real robotic systems}
The sensor was applied in two distinct robotic case studies: guiding a robot by hand (using tactile commands) and human-robot collision avoidance (measuring hand proximity). Owing to its flexibility, the 16 prexel sensor was installed around the robot end-effector as if it was masking tape, Fig. \ref{fig:fig12}. The collaborative robot is a 7-DOF KUKA iiwa R800 equipped with the Sunrise controller and interfaced using the KST Toolbox. Data were acquired from the sensor at 10 Hz for proximity readings, and 100 Hz for tactile robot control.

\subsection{Robot Hand Guidance}

For the robot hand-guiding application, force-sensing measurements are acquired, treated and transformed into robot motion commands, i.e., translations in Cartesian space. Proximity sensing was also used to trigger the robot hand-guiding control mode. 

\begin{figure}[!t]\centering
	\includegraphics[width=1\columnwidth]{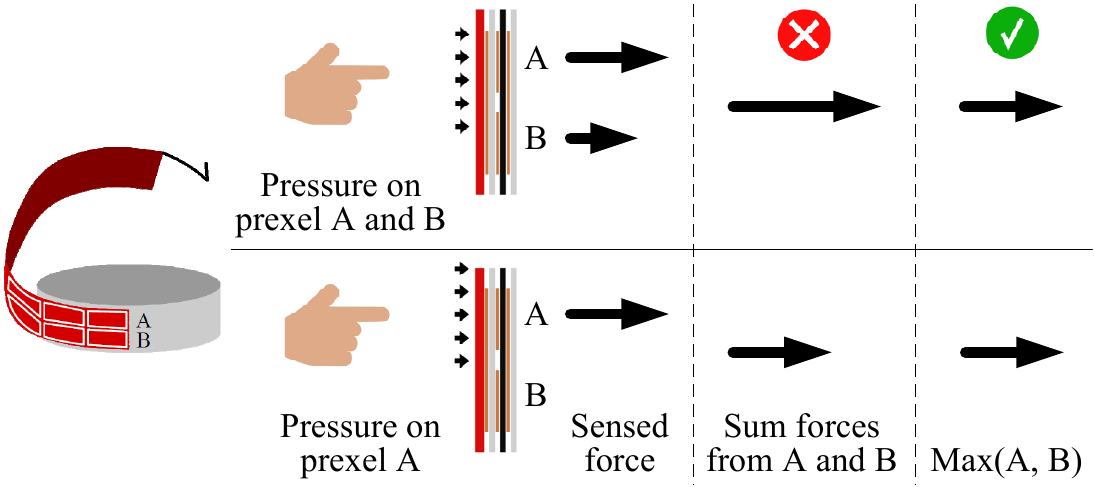}\centering
	
	\caption{For x-y axis robot commands, only the maximum forces sensed across all lines of each column of the FSR array are considered, to avoid accidental overmeasures.} \label{fig:fig12}
\end{figure}

The 16 prexel sensor variant features a 2 $\times$ 8 FSR array with 2 rows (A and B) and 8 columns (1, 2, ..., 8), Fig. \ref{fig:fig12}. Each column of the array, when touched, triggers a corresponding robot motion in x-y Cartesian space. Each finger touching the sensor spreads the pressure between one or both rows in the array (A and B), Fig. \ref{fig:fig12}. Only the maximum force, applied to prexel (A, $n$) or prexel (B, $n$), is considered. Adding forces sensed across both rows would result in overmeasures. The sensor\textquotesingle s width (15 mm) ensures that at least one of the two prexels in each column is fully covered by the human finger. To hand guide the robot along the z-axis, we calculate the pressure difference between both rows of prexels. The human finger tends to touch the sensor in the centre (the sensor's 15 mm width forces that scenario), leading to an approximately equal pressure distribution between FSR array rows. When the hand tries to push the robot upwards, or downwards, an angular momentum is created between the skin's surface (where a traction force is applied) and the bone inside. This momentum changes the balance of pressures, which can be calculated and used to indicate the human\textquotesingle s intention to move the robot end-effector along the z-axis. The sensor is calibrated by attaching a reference frame to its structure (matrix of prexels whose position is fixed and known). The relative position of each prexel is calculated in relation to the robot\textquotesingle s local reference frame using homogeneous transformations. The sensor is sticked around the robot\textquotesingle s tool flange in a known position and orientation relative to the robot\textquotesingle s end-effector reference frame, Fig. \ref{fig:fig15} b).

\begin{figure}[!t]\centering
	\includegraphics[width=1\columnwidth]{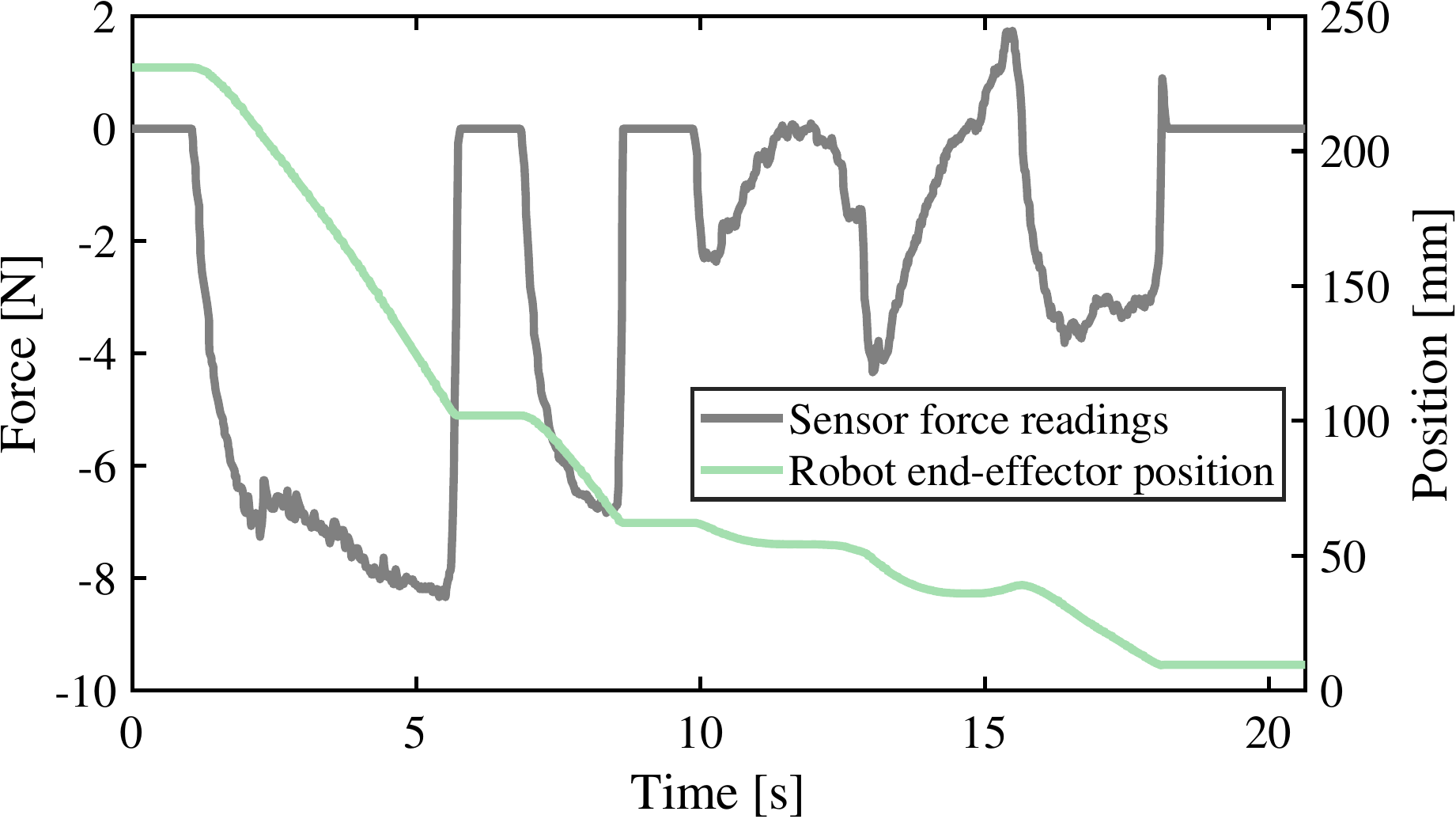}\centering
	
	\caption{Data recorded during robot hand guiding demonstration. Plotted forces and end-effector position are both relative to the x-axis direction, being the robot's response analogous for the y-axis and z-axis directions.} \label{fig:fig13}
\end{figure}

The experimental test is shown in time-lapse, Fig. \ref{fig:fig15} b). The robot end-effector position along the x-axis is plotted against sensor force measurements in Fig. \ref{fig:fig13}. This analysis along a single axis, x-axis in this case, facilitates the visualization and understanding of the system\textquotesingle s behaviour which can be extrapolated to the other axes. 

Five different users indicated they found the hand-guiding process to be more intuitive than the traditional teach-in robot programming (using the teach pendant). Three users are researchers and two are project engineers, all of them with basic skills in robotics. They were informed that the robot\textquotesingle s end-effector could be moved by simply pushing on its tool flange (where the sensor was installed), without further explanation. As the robot started to move, all users intuitively pushed or pulled the robot flange in different directions, immediately and intuitively realising how the system behaves/operates.

\begin{figure}[!t]\centering
	\includegraphics[width=1\columnwidth]{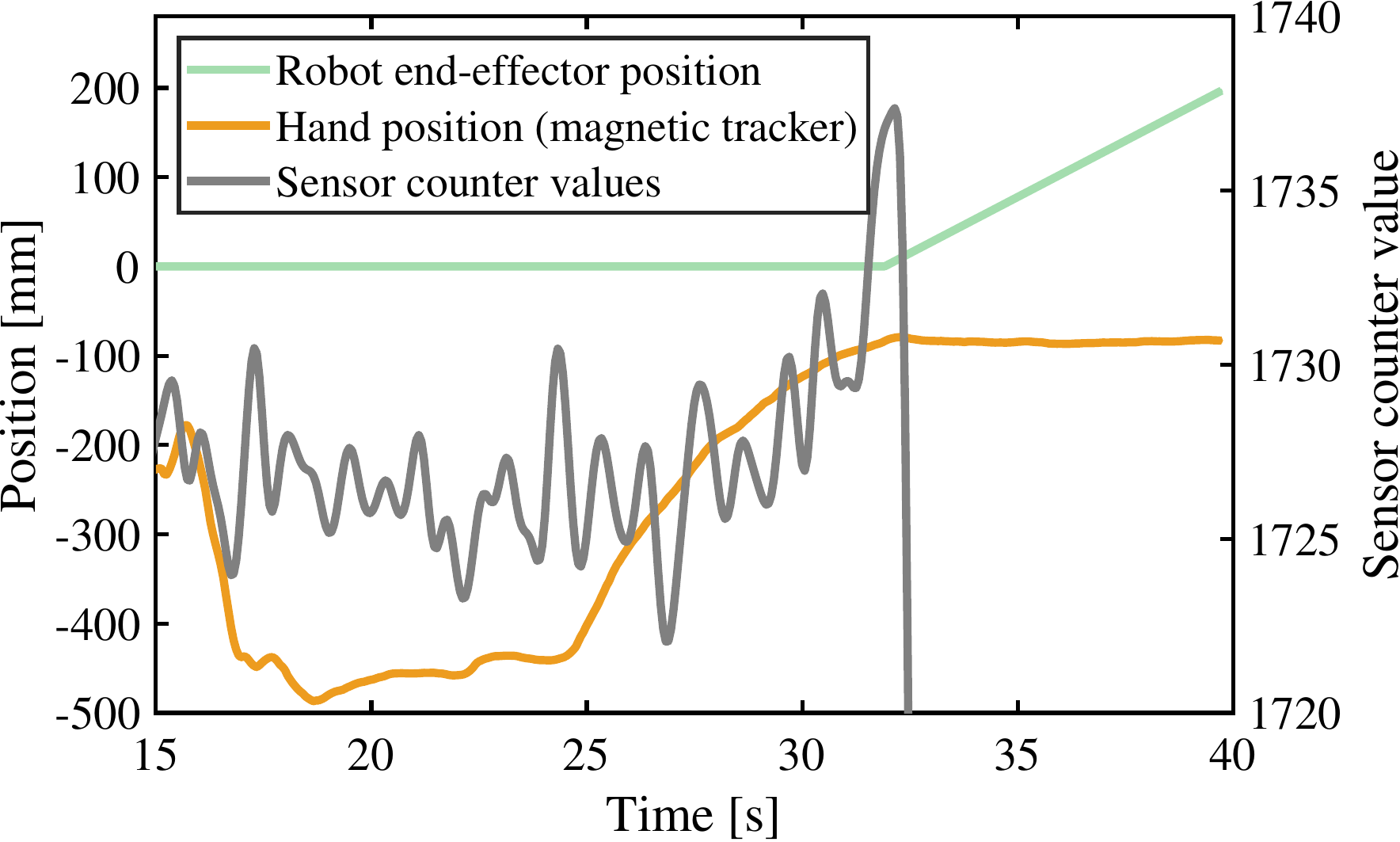}\centering
	
	\caption{Human-robot collision avoidance demonstration. The 64 prexel sensor triggers an emergency command to the robot when a human hand reaches within 100 mm of the robot's end-effector. All data are unfiltered.} \label{fig:fig14}
\end{figure}

\begin{figure*}\centering
	\includegraphics[width=1\textwidth]{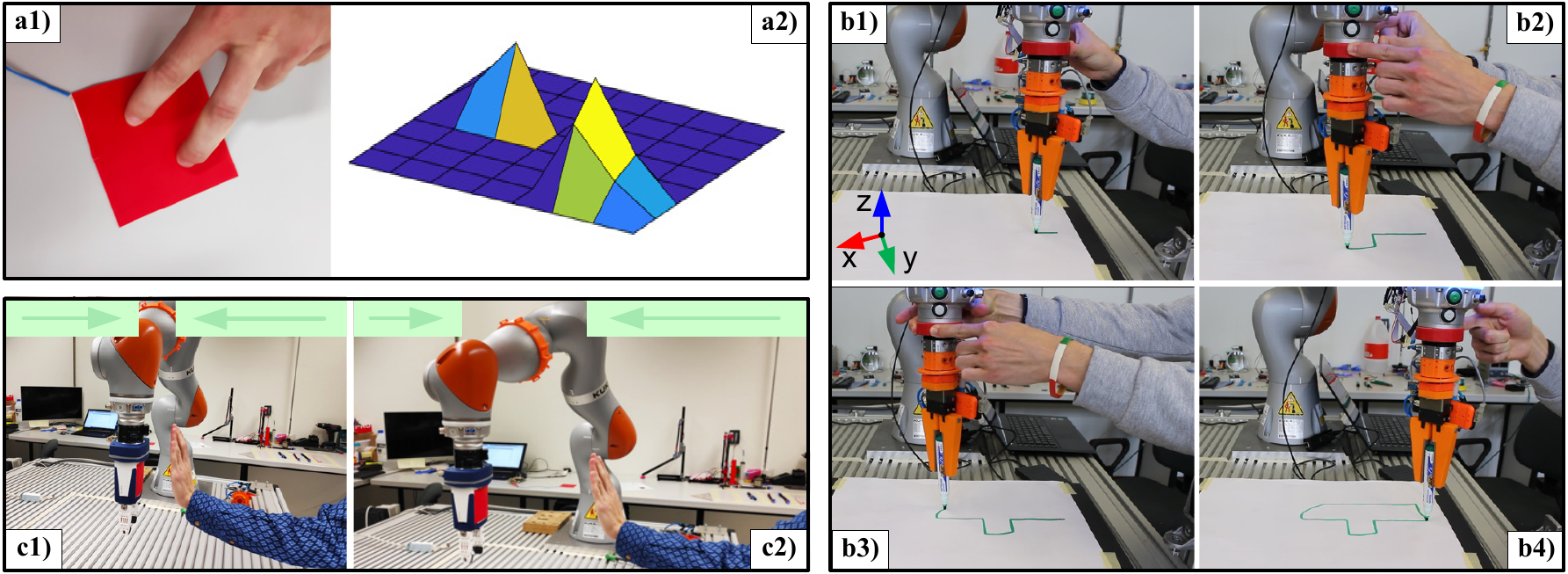}\centering
	\caption{a) Two fingers touching the 64 prexel sensor (left) and corresponding response plot (right), b) Robot hand-guiding with the smaller 16 prexel sensor installed around the robot's end-effector, c) Human-robot collision avoidance test with the 64 prexel sensor: an emergency command is triggered when the hand reaches within 100 mm from sensor. The robot moves to a safe position to avoid collisions.}
	\label{fig:fig15}
\end{figure*}

\begin{figure}[!t]\centering
	\includegraphics[width=1\columnwidth]{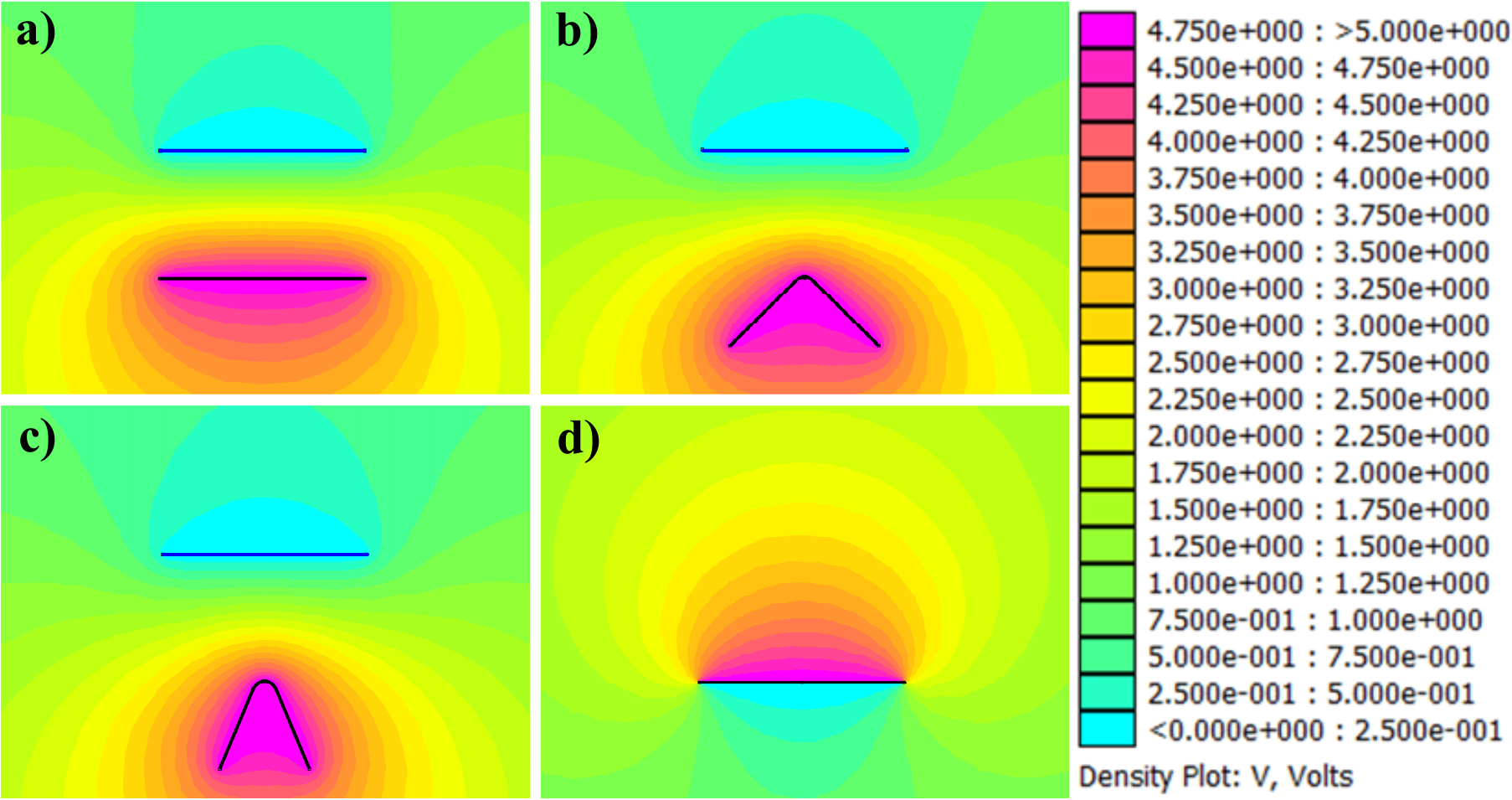}\centering
	
	\caption{Electrostatic potential around 8 cm$^2$ self-capacitive electrode with 8 cm$^2$ test grounded electrode in close proximity (a, b and c). d) Electrostatic potential around 8 $\times$ 8 sensor with FSR layers being used as active electrostatic shields.} \label{fig:fig16}
\end{figure}

\begin{figure}[!t]\centering
	\includegraphics[width=1\columnwidth]{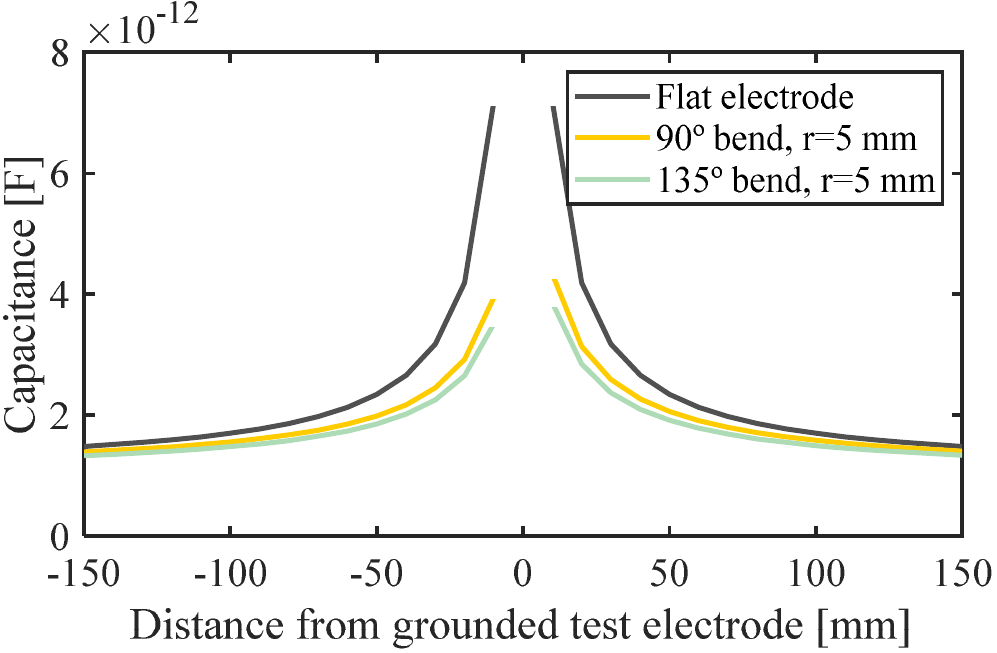}\centering
	
	\caption{Capacitance measured in 8 cm$^2$ electrode.} \label{fig:fig17}
\end{figure}

\subsection{Human-Robot Collision Avoidance}

In order to avoid human-robot collisions, the hand proximity to the robot was monitored using the sensor\textquotesingle s self-capacitive operational mode. For this specific experimental test, the sensor was attached to the robot gripper in a relatively flat surface. Experiments demonstrated that it can be applied in various locations of the robot arm, in both flat and moderately curved surfaces, keeping an efficient monitoring. Fig. \ref{fig:fig15} c) shows that when the sensor detects a human hand in close proximity, a command is issued to the robot controller, immediately moving the robot to a safe, pre-determined position. The system latency is approximately 100 ms and the maximum human-robot relative speed during the test was 60 mm/s. Fig. \ref{fig:fig14} shows the distance between the hand and the robot, measured by both the proposed sensor and an electromagnetic tracker (Polhemus Liberty), as well as the robot end-effector position. All data are unfiltered.

Results demonstrate that the sensor provides human-robot relative proximity data and by this way the robot reacts to avoid collision.

\subsection{Sensor Robustness and Shortcomings}

The sensor showed significant deviations (over 50\%) on proximity measurements, when placed at less than 100 mm from the robot\textquotesingle s actuators (while the robot was moving). Preliminary simulations show a vast reduction electrostatic potential behind the shielded side of the 8 $\times$ 8 sensor, Fig. \ref{fig:fig16} d). This is expected to increase the sensor\textquotesingle s robustness to electrostatic interference originated from behind it. Although implementation of this feature requires no modification to the sensor, further development of the DAQ circuit will be required. The current circuit is unable to simultaneously address all the FSR electrodes, a feature which would be necessary to use them as active shields.

Stress concentration and pre-stress may occur when the sensor is applied to curved surfaces. The sensor was mounted on a 30 mm constant-radius curved surface. In this case, we simply tared the sensor and achieved results within the same $\sim$10\% error margin obtained over a flat surface.  Creasing prexels around sharp edges should be avoided where possible to prevent stress concentration in the piezoresistive polymer. In cases where this is not feasible, local calibration of the creased prexels must be performed, and a decrease in maximum measurable force is to be expected due to the increased stress resultant from the reduced contact area between the object/human and the edge on which the sensor is applied. When applied to complex geometry surfaces, the normal distance between a hand within proximity range and the sensor varies according to surface topography, causing self-capacitance measurements to fluctuate accordingly. This effect can be simulated using the method of moments for electrostatics \cite{ref26}. To illustrate this point, capacitance on the 8 cm$^2$ electrode was calculated when another 8 cm$^2$ grounded conductor is placed within proximity range, simulating a hand at distances ranging from 10 mm up to 150 mm, Fig. \ref{fig:fig16} a). The same simulation was then repeated with the self-capacitive electrode bent 90$^\circ$ and 135$^\circ$ around a 5 mm curvature radius, Fig. \ref{fig:fig16} b) and Fig. \ref{fig:fig16} c), respectively. Results are presented in Fig. \ref{fig:fig17}, where the capacitance reduction with increasing curvature becomes clear. In practical terms, this results in the sensor not being capable of accurately determining hand-to-surface distance when applied to highly curved surfaces, being however still moderately effective at detecting hand presence within its proximity range.

\section{Conclusion and Future Work}
This paper has presented a novel mechanically flexible piezoresistive/self-capacitive hybrid force and proximity sensor, which proved to be multifunctional, easy to fabricate, highly inexpensive and easy to apply to complex-shaped robot structures. These characteristics demonstrated its potential application in collaborative robot interfaces. Experimental results, for both the piezoresistive and self-capacitive operation modes, showed the sensor's versatility, flexibility (1 mm thickness), accuracy (reduced drift) and repeatability. Its fabrication method is simple and highly scalable while being flexible enough to accept process variations. Finally, the sensor was successfully applied in real-world scenarios, namely robot hand guiding and human-robot collision avoidance. Future work will be dedicated to testing 3D printer-based manufacturing processes based on conductive filaments and improvement of the sensor\textquotesingle s proximity range and robustness to electromagnetic interference by utilizing the aforementioned FSR electrodes as active electrostatic shields during proximity measurement cycles.


\begin{thebibliography}{00}

\bibitem{ref1}
Guang-Zhong Yang and et~al.
\newblock The grand challenges of science robotics.
\newblock {\em Science Robotics}, 3(14), 2018.

\bibitem{ref2}
Jeremy~A. Marvel.
\newblock Sensors for safe, collaborative robots in smart manufacturing.
\newblock In {\em 2017 IEEE SENSORS}, pages 1--3, 2017.

\bibitem{ref3}
Jos{\'e} Saenz, R.~Behrens, E.~Schulenburg, Hauke Petersen, O.~Gibaru, P.~Neto,
and N.~Elkmann.
\newblock Methods for considering safety in design of robotics applications
featuring human-robot collaboration.
\newblock {\em The Int J of Advanced Manufacturing Technology}, 107:2313--2331,
2020.

\bibitem{ref4}
Hyo~Seung Han, Junwoo Park, Tien~Dat Nguyen, Uikyum Kim, Soon~Cheol Jeong,
Doo~In Kang, and Hyouk~Ryeol Choi.
\newblock {A flexible dual mode tactile and proximity sensor using carbon
	microcoils}.
\newblock In Yoseph Bar-Cohen and Frédéric Vidal, editors, {\em Electroactive
	Polymer Actuators and Devices (EAPAD) 2016}, volume 9798, pages 634 -- 640.
International Society for Optics and Photonics, SPIE, 2016.

\bibitem{ref5}
Tien~Dat Nguyen, Taesung Kim, Jiho Noh, Hoa Phung, Gitae Kang, and Hyouk~Ryeol
Choi.
\newblock Skin-type proximity sensor by using the change of electromagnetic
field.
\newblock {\em IEEE Transactions on Industrial Electronics}, 68(3):2379--2388,
2021.

\bibitem{ref6}
Lucie Viry, Alessandro Levi, Massimo Totaro, Alessio Mondini, Virgilio Mattoli,
Barbara Mazzolai, and Lucia Beccai.
\newblock Flexible three-axial force sensor for soft and highly sensitive
artificial touch.
\newblock {\em Adv Materials}, 26(17):2659--2664, 2014.

\bibitem{ref7}
Julian Castellanos~Ramos, Andres Trujillo~Leon, Rafael Navas~Gonzalez,
Francisco Barbero~Recio, Jose~Antonio Sanchez-Duran, Óscar Oballe~Peinado,
and Fernando Vidal~Verdu.
\newblock Adding proximity sensing capability to tactile array based on
off-the-shelf fsr and psoc.
\newblock {\em IEEE Transactions on Instrumentation and Measurement},
69(7):4238--4250, 2020.

\bibitem{ref8}
Luheng Wang.
\newblock Usage of connected structure to eliminate blind area of
piezoresistive sensor array.
\newblock {\em IEEE Transactions on Industrial Electronics}, 65(4):3568--3575,
2018.

\bibitem{ref9}
Yuanzhao Wu, Yiwei Liu, Youlin Zhou, Qikui Man, Chao Hu, Waqas Asghar, Fali Li,
Zhe Yu, Jie Shang, Gang Liu, Meiyong Liao, and Run-Wei Li.
\newblock A skin-inspired tactile sensor for smart prosthetics.
\newblock {\em Science Robotics}, 3(22), 2018.

\bibitem{ref10}
Takahiro Matsuno, Zhongkui Wang, Kaspar Althoefer, and Shinichi Hirai.
\newblock Adaptive update of reference capacitances in conductive fabric based
robotic skin.
\newblock {\em IEEE Robotics and Automation Letters}, 4(2):2212--2219, 2019.

\bibitem{ref11}
Josie Hughes and Fumiya Iida.
\newblock Tactile sensing applied to the universal gripper using conductive
thermoplastic elastomer.
\newblock {\em Soft Robotics}, 5(5):512--526, 2018.

\bibitem{ref12}
Naoji Matsuhisa, Martin Kaltenbrunner, Tomoyuki Yokota, Hiroaki Jinno, Kazunori
Kuribara, Tsuyoshi Sekitani, and Takao Someya.
\newblock Printable elastic conductors with a high conductivity for electronic
textile applications.
\newblock {\em Nature Communications}, 6(1):7461, Jun 2015.

\bibitem{ref13}
Artem Dementyev, Hsin-Liu~(Cindy) Kao, and Joseph~A. Paradiso.
\newblock Sensortape: Modular and programmable 3d-aware dense sensor network on
a tape.
\newblock In {\em Proceedings of the 28th Annual ACM Symposium on User
	Interface Software \& Technology}, UIST '15, pages 649--658, New York, NY,
USA, 2015. Association for Computing Machinery.

\bibitem{ref14}
Dana Hughes, John Lammie, and Nikolaus Correll.
\newblock A robotic skin for collision avoidance and affective touch
recognition.
\newblock {\em IEEE Robotics and Automation Letters}, 3(3):1386--1393, 2018.

\bibitem{ref15}
Emmanuel Dean-Leon, Karinne Ramirez-Amaro, Florian Bergner, Ilya Dianov, and
Gordon Cheng.
\newblock Integration of robotic technologies for rapidly deployable robots.
\newblock {\em IEEE Transactions on Industrial Informatics}, 14(4):1691--1700,
2018.


\bibitem{ref16}
Junwen Zhong, Zhaoyang Li, Masahito Takakuwa, Daishi Inoue, Daisuke Hashizume,
Zhi Jiang, Yujun Shi, Lexiang Ou, Md~Osman~Goni Nayeem, Shinjiro Umezu,
Kenjiro Fukuda, and Takao Someya.
\newblock Smart face mask based on an ultrathin pressure sensor for wireless
monitoring of breath conditions.
\newblock {\em Advanced Materials}, 34(6):2107758, 2022.

\bibitem{ref17}
Sunghoon Lee, Sae Franklin, Faezeh~Arab Hassani, Tomoyuki Yokota, Md~Osman~Goni
Nayeem, Yan Wang, Raz Leib, Gordon Cheng, David~W. Franklin, and Takao
Someya.
\newblock Nanomesh pressure sensor for monitoring finger manipulation without
sensory interference.
\newblock {\em Science}, 370(6519):966--970, 2020.

\bibitem{ref18}
Sungwon Lee, Amir Reuveny, Jonathan Reeder, Sunghoon Lee, Hanbit Jin, Qihan
Liu, Tomoyuki Yokota, Tsuyoshi Sekitani, Takashi Isoyama, Yusuke Abe, Zhigang
Suo, and Takao Someya.
\newblock A transparent bending-insensitive pressure sensor.
\newblock {\em Nature Nanotechnology}, 11(5):472--478, May 2016.

\bibitem{ref19}
Takao Someya and Tsuyoshi Sekitani.
\newblock Printed skin-like large-area flexible sensors and actuators.
\newblock {\em Procedia Chemistry}, 1(1):9--12, 2009.
\newblock Proceedings of the Eurosensors XXIII conference.


\bibitem{ref20}
D.~Giovanelli and E.~Farella.
\newblock Force sensing resistor and evaluation of technology for wearable body
pressure sensing.
\newblock {\em J. Sensors}, 2016, 2016.


\bibitem{ref21}
Luheng Wang, Tian Ding, and Peng Wang.
\newblock Influence of carbon black concentration on piezoresistivity for
carbon-black-filled silicone rubber composite.
\newblock {\em Carbon}, 47:3151--3157, 2009.


\bibitem{ref22}
J.~G. Simmons.
\newblock Generalized formula for the electric tunnel effect between similar
electrodes separated by a thin insulating film.
\newblock {\em Journal of Applied Physics}, 34(6):1793--1803, 1963.

\bibitem{ref23}
Paul Bagder and Paul Stoffregen.
\newblock Capacitive sensing library.

\bibitem{ref24}
Mohammad Safeea and Pedro Neto.
\newblock Kuka sunrise toolbox: Interfacing collaborative robots with matlab.
\newblock {\em IEEE Robotics \& Automation Magazine}, 26(1):91--96, 2019.

\bibitem{ref25}
Masoud Kalantari, Javad Dargahi, Jozsef K\"{o}vecses, Mahmood~Ghanbari Mardasi,
and Shahrzad Nouri.
\newblock A new approach for modeling piezoresistive force sensors based on
semiconductive polymer composites.
\newblock {\em IEEE/ASME Transactions on Mechatronics}, 17(3):572--581, 2012.

\bibitem{ref26}
T.K. Sarkar and R.F. Harrington.
\newblock The electrostatic field of conducting bodies in multiple dielectric
media.
\newblock {\em IEEE Transactions on Microwave Theory and Techniques},
32(11):1441--1448, 1984.




\end{thebibliography}
\end{document}